\documentclass[fleqn,10pt]{wlpeerj} 
\usepackage{natbib}
\usepackage{float}
\usepackage{xcolor}

\title{Evaluating and explaining training strategies for zero-shot cross-lingual news sentiment analysis}

\author[1]{Luka Andrenšek\footnote{equal contirbution}}
\author[1,2]{Boshko Koloski\textsuperscript{*}}
\author[1]{Andraž Pelicon}

\author[1]{Nada Lavrač}

\author[1]{Senja Pollak}
\author[1,3]{Matthew Purver}

\affil[1]{Jožef Stefan Institute, Ljubljana, Slovenia}
\affil[2]{Jožef Stefan International Postgraduate School, Ljubljana, Slovenia}
\affil[3]{Queen Mary University of London, London, United Kingdom}
\corrauthor[1]{Boshko Koloski}{boshko.koloski@ijs.si}

\begin{abstract}
We investigate  zero-shot cross-lingual news sentiment detection, aiming to develop robust sentiment classifiers that can be deployed across multiple languages without target-language training data. We introduce novel evaluation datasets in several less-resourced languages, and experiment with a range of approaches including the use of machine translation; in-context learning with large language models; and various intermediate training regimes including a novel task objective, POA, that leverages paragraph-level information. Our results demonstrate significant improvements over the state of the art, with in-context learning generally giving the best performance, but with the novel POA  approach giving a competitive alternative with much lower computational overhead. We also show that language similarity is not in itself sufficient for predicting the success of cross-lingual transfer, but that similarity in semantic content and structure can be equally important.
\end{abstract}

\begin{document}

\flushbottom
\maketitle
\thispagestyle{empty}

\section*{Introduction}

Sentiment analysis (SA) --- identifying the sentiment of textual documents --- is one of the most popular natural language processing (NLP) tasks \cite{goddard2011semantic}. SA can be formulated to analyze the sentiment from the readers' perspective, the sentiment of the writer towards particular subject and so on;  has been applied to a variety of data sources, including social media posts \cite{socmedi1231}, news articles \cite{app10175993}, reviews \cite{Fang2015}  and similar; %Given the level of analysis of a document 
and can be performed at the sentence, the paragraph or the document level \cite{medhat2014sentiment}. The methods can be either monolingual (training and testing on the same language) or cross-lingual (training on a \emph{source} language and testing on a different \emph{target} language), but in either case it is generally approached as a supervised learning problem; this imposes a need for labeled data, seen as one of the main obstacles in developing robust systems for NLP, and especially difficult with low-resource languages. For this reason,
research has been focused lately on models that can work cross-lingually and preferably in a zero-shot setting (requiring no training in the target language). We are especially interested in news sentiment analysis, and in building cross-lingual models, allowing for SA %sentiment analysis 
of  news document collections in several languages. 

We recognise several variants of %document-level 
cross-lingual SA, %sentiment analysis 
%in cross-lingual scenario 
based on the %data
availability of %the
target language data:\textit{ zero-shot} (when no target data is %present
available for training), \textit{few-shot} (where only about a dozen instances are available) and \textit{full-shot }(where %bigger 
larger amounts of data are available). % in the target language). 
This paper focuses on the zero-shot cross-lingual setting, where models trained on available data in one language are applied to predict sentiment in other language without any additional training on the target language. %, just by providing a few example pairs to the model \citep{} \mpu{missing ref?}. This strategy is particularly promising for less-resourced languages, which often lack the %vast %\mpu{try to avoid "vast", it sounds amateurish}
%large amounts of annotated data required to train robust models. %Zero-shot learning is efficient because it leverages the generalizability of models trained on rich-resource languages, allowing them to understand and process languages with limited computational resources.
By using this approach, we can extend the benefits of advanced NLP techniques to a broader range of languages, promoting linguistic inclusivity and enhancing the understanding of sentiment across diverse linguistic landscapes. \citet{app10175993} investigated the zero-shot cross-lingual capabilities of the mBERT \cite{devlin-etal-2019-bert} model for news SA, while the same paradigm has been applied to may other tasks, including cross-lingual document retrieval \cite{funaki2015image}, sequence labeling \cite{rei2018zero,koloski-etal-2022-thin}, cross-lingual dependency parsing \cite{wang2019cross} and reading comprehension \cite{hsu2019zero}. %Lausc her et al. \mpu{agh, stop it}
\citet{lauscher-etal-2020-zero} pinpoint the limited capabilities of zero-shot models and propose few-shot learning when possible. 

While news sentiment has been explored in various studies \cite{balahur2013sentiment}, the zero-shot perspective is rarely taken into account \cite{app10175993}, and specifically, South-Slavic languages are rarely included in evaluation.
In our study, we build upon the work of %Pelicon et al. 
\citet{app10175993}, who investigated the transfer of news SA %sentiment 
from Slovenian to Croatian, but extend the cross-lingual news sentiment evaluation to new languages with several novel evaluation datasets, and compare various training strategies. We assess the performance of models on Croatian, Albanian, Estonian, Bosnian, Serbian, and Macedonian datasets, while training the model itself %was trained 
on a dataset of Slovenian news articles.

We extend the approach of \citet{app10175993} in several ways. First, we investigate the \emph{in-context learning} approach \cite{dong2023survey},
using a Large Language Model (LLM) and passing labelled examples directly within the input data prompt at test time, rather than through a separate training phase. This method leverages patterns and relationships inherently present in the data to make predictions or generate responses. By integrating examples into the input, the model dynamically adapts to the context of the new data.  This enables us to utilise the model for downstream classification in the target language without any need for target language data. % With this approach, we avoid the need for computationally intensive re-training? and overcome the limitations of a fixed size context window, as the LLM can accept input length, greater than any length of our documents.\mpu{not quite clear what we mean here: if just that our LLM has a fixed window size but it's big enough for our data, that's not really "overcoming the limitations" of fixed size, it's just being lucky that the limitations don't happen to limit us in this case; if it's that you apply the LLMs as language models in a way that there is no window, need to say that more clearly}
Secondly, we introduce a novel approach to facilitate cross-lingual transfer, specifically introducing
a novel Part Of Article (POA) approach. This approach not only informs the model about the sentiment of a text segment but also its relative location within the document, giving the model more information about how a particular segment contributes to the overall sentiment of the document. Additionally, we couple this with an intermediate sentiment enrichment step introduced in  \cite{app10175993}, an extended in-domain pretraining step combining the Masked Language objective and paragraph level sentiment classification. We also systematically compare the effect of using translation to English or keeping the original language; as English is better covered in pretrained models, we hypothesise it can improve the performance.

The main contributions of our paper are: i) novel news sentiment annotations for Macedonian, Serbian, Albanian, Bosnian; ii) a novel Part Of Article (POA) approach, facilitating this cross-lingual sentiment transfer by paragraph position information; iii) systematic comparison of a range of cross-lingual SA approaches, including translating the entire corpus to a well resourced language; enriching the model with additional (monolingual) domain-relevant sentiment information; enriching the article with relative position information (POA); and in-context learning in both multilingual and translated monolingual settings and iv) a semantic approach towards explanation of transferability gap between perfromance in source and target languages.

% \mpu{yes, it is, but basically all papers are, so no need to say this. And it's not really useful for the reader anyway}
%Our study is organized into sections detailing related works, the datasets used, our methodology, and the experimental setup. 
% \mpu{and yes, we do, but all authors do, so see above etc etc}
%We discuss our findings.

\section{Related work}
\label{sec:related_work}

%In this section, we provide an overview over the different approaches used for sentiment analysis.  
While early approaches towards sentiment analysis were based on handcrafted sentiment lexicons like VADER \cite{hutto2014vader}, most recent work uses machine learning due to its higher accuracy, robustness and ability to transfer to new domains and languages without needing to create new resources. In this section, we give an overview of the commonly used traditional machine learning approaches, deep learning architectures and finally the more recent pre-trained large language model based approaches.

%\subsection{Traditional approaches} While early approaches towards sentiment analysis based on preconstructed sentiment lexicons like VADER \cite{hutto2014vader}, most recent work uses machine learning due to its higher accuracy, robustness and ability to transfer to new domains and languages without needing to create new resources.   
%Sentiment analysis (SA) has traditionally been anchored in machine learning, \mpu{there are lots of non-ML approaches e.g. lexicon-based ones like VADER, and you could argue they're even more "traditional" really. So maybe we need to start by justifying our ML approach just briefly, e.g. "While some work in SA uses fixed lexicons, manually defined or automatically extracted [cite VADER, SentiNet etc], most recent work uses machine learning due to its higher accuracy, robustness and ability to transfer to new domains and languages without needing to create new resources."} 

\subsection{Machine learning} Machine learning-based approaches \cite{zhang2016comparison} for SA are often based on two constructs. The first consists of featurizing the input text, usually constructing interpretable features from the bag-of-words family:  word and/or character n-gram counts, lexical features, sentiment lexicon-based features, parts of speech, or adjectives and adverbs, often weighting these via %some sophisticated counting scheme like 
e.g.\ term frequency-inverse document frequency \cite{ref1}. The second part is based on learning a classifier on top of these features. \citet{mehra2002sentiment} adopted a maximum entropy classifier, while \citet{maliknb} utilised a Naive Bayes classifier. A support vector machine (SVM) classifier for SA on tweets was explored by \citet{firmino2014comparison}, based on linear classifiers such as SVMs and logistic regression \cite{pang-etal-2002-thumbs}. These methods have established a robust baseline for the field. However, they suffer from significant drawbacks. First, they are based on counting and %completely 
ignore any deep semantics or word order. Next, they generalise poorly out-of-distribution, let alone across languages, due to the fact that in zero-shot cross-lingual applications, the vocabularies can be completely different. Finally, the classifiers are trained on a large number of features, which \citet{hughes} argues negatively impacts the classifier, resulting in lower classifier accuracy. To overcome some of these issues, researchers next explored deep learning based approaches \cite{lecun2015deep}.
%With the advancement of deep learning, newer models, particularly those based on neural networks, have demonstrated superior performance due to their ability to capture contextual dependencies in text \cite{socher-etal-2013-recursive, zhang2018character}. 

\subsection{Deep learning approaches} The deep learning approaches are based on the idea that both the representation and the classification functions can be learned end-to-end in a differentiable fashion. \citet{glorot2011domain} proposed a deep learning-based approach towards sentiment analysis across different domains. The core foundation of this work was that representations for the words and documents can be learned from the data in an unsupervised fashion, and then a classifier can exploit these semantically rich features. In a similar vein, \citet{socher2011semi} proposed learning recursive autoencoders in a semi-supervised setting for SA, which outperformed the traditional machine learning approach. Variational autoencoders based on Long-Short Term Memory (LSTM) networks \cite{WU201930} have also been proposed to tackle this task successfully. The shift towards deep learning for SA is marked by the wide adoption of convolutional neural networks (CNNs) and recurrent neural networks (RNNs). These models benefit from the deep architectures' ability to learn hierarchical representations of text data, significantly enhancing nuance and accuracy. Efforts to improve both monolingual and cross-lingual sentiment analysis have led to various flavours of RNNs and CNNs being explored ~\cite{dong2018cross,abdalla2017cross,ghasemi2022deep,baliyan2021multilingual,kanclerz2020cross}.

\subsection{Pretrained models}
The use of pre-trained Transformer models, such as BERT (Bidirectional Encoder Representations from Transformers) and its multilingual derivatives, mBERT \cite{devlin-etal-2019-bert} and XLMR \cite{conneau-etal-2020-unsupervised}, has set new standards in text classification in both monolingual and cross-lingual settings. Despite their widespread adoption, these models face limitations \cite{peng2024limitations} like fixed input lengths and a potential lack of sensitivity to sentiment nuances due to their pre-training focus on linguistic structures rather than emotional content. To address the structural lack of emotional content, \citet{yin-etal-2020-sentibert} proposed the SentiBERT model, capable of capturing semantic composition by synergising contextualised representations with binary constituency parse trees. In another study, \citet{ke-etal-2020-sentilare} enhanced pre-trained transformer representations with word-level linguistic knowledge, such as part-of-speech tags and sentiment polarity, via label-aware masked-language modelling, further improving the representations.

Although these improvements in pre-trained representations are significant, no general approach exists for multilingual sentiment detection. \citet{app10175993} investigated how training a multilingual model to analyse sentiment in one language could be leveraged for zero-shot applications such as news sentiment analysis. Similarly, \citet{robniksikonja2021crosslingual} explored cross-lingual transfer. The introduction of cross-lingual pre-training \cite{conneau2019cross} enabled the creation of powerful multilingual models capable of cross-lingual transfer. Models pre-trained with multilingual masked-language modelling \cite{conneau-etal-2020-unsupervised} exhibit strong cross-lingual understanding due to the compositional nature of language, as demonstrated by \cite{chai-etal-2022-cross}.

Core challenges in the cross-lingual sentiment detection field are surveyed in \cite{Xu2022}, which extends to maintaining sentiment accuracy across languages without extensive labelled data in each target language. Early strategies often involved direct translation of text data followed by sentiment analysis using a model trained in the source language \cite{banea-etal-2008-multilingual}. The rise of neural machine translation has facilitated more sophisticated approaches, such as training joint bilingual sentiment models that leverage shared representations across languages \cite{chen-cardie-2018-multinomial}. Various methodologies have been developed to leverage machine translation and transfer learning techniques to overcome language barriers in NLP tasks.

Among these, the T3L: Translate-and-Test Transfer Learning for Cross-Lingual Text Classification introduced by \citeauthor{unanue2023t3l} is a notable contribution \cite{unanue2023t3l}. T3L capitalises on a translate-and-test strategy that simplifies adapting models for multiple languages by enabling end-to-end fine-tuning with soft parameter sharing. While their approach introduced heavier memory restrictions and slower inference speeds, it achieved competitive text classification results. Another significant approach is by \citeauthor{siddhant2020crosslingual}, who evaluated the cross-lingual effectiveness of massively multilingual neural machine translation \cite{siddhant2020crosslingual}, which included 103 languages. Their findings underscore the potential of using a unified model architecture to perform NLP tasks across multiple languages, thereby reducing the overhead of maintaining separate models for each language while still maintaining performance across multiple NLP tasks comparable to specialised approaches.

Additionally, \citeauthor{ren2019explicit}'s work on explicit cross-lingual pre-training for unsupervised machine translation provides an insightful approach to enriching models with additional cross-lingual information, which is crucial for languages with limited annotated resources \cite{ren2019explicit}. Incorporating explicit cross-lingual training signals enhances a model's performance on multiple tasks. Recent advances also include the use of multilingual transformers such as mBERT and XLMR, which provide pre-trained models capable of understanding multiple languages, thereby allowing for more effective transfer learning applications in sentiment analysis across different linguistic datasets \cite{conneau-etal-2020-unsupervised}. These models have significantly lowered the barriers to entry for cross-lingual NLP tasks by providing a robust pre-trained foundation that can be fine-tuned on relatively small amounts of task-specific data. However, these models do not receive information conveying sentiment in their pre-training phase, opening possibilities of making these models more sensitive to sentiment information by introducing an additional pre-training step that enriches them with contextual and sentiment-sensitive information, as shown by \citet{app10175993}.

With the introduction of the NLLB (No Language Left Behind) model \cite{nllbteam2022language}, capable of translating between more than 200 pairs of languages, a significant language barrier in NLP tasks has been mitigated. The traditional approach of translation followed by classification has opened new possibilities for achieving competitive performance results.

%\mpu{This section needs a bit more structure, at the moment it feels a bit like just a list of short paragraphs, each one describing a paper, but without much of a story. We could explain the need for cross-lingual / zero-shot methods here, for example.}

\section{Data}

In this Section~\ref{seubs:data_dsc}, we provide preliminary statistics for the datasets used. The process of constructing these datasets is detailed in Section~\ref{seubs:data_creation}. Upon acceptance, we will publish the new datasets.

\subsection{Data description}
\label{seubs:data_dsc}

We work with seven datasets, each containing general news articles in a different language. Each document is labelled as \textit{negative}, \textit{neutral} or \textit{positive}.  The languages, with document counts, %and the number of documents in each 
are as follows: Slovenian (10,427 documents), Croatian (2,025), Bosnian (200), Macedonian (198), Estonian (100), Serbian (200), and Albanian (200); see Table~\ref{tab:data}. %\mpu{maybe nicer to order the last 3 columns pos-neu-neg (rather than pos-neg-neu)?}

\begin{table}[H]
    \centering
\resizebox{\textwidth}{!}{\begin{tabular}{lrrrrrr|rrr}
\toprule
%language & Num. Doc. & Doc. Len. & Avg. Num. Words & Avg. Word Len. & Avg. Sent. Len. & Num. Sent. & positive & negative & neutral \\
%MP: adjusting to allow the number font size to be bigger in the table
language & \# Doc. & Doc. Len. & Avg. \#   & Avg. Word  & Avg. Sent.  & \#  Sent. & positive &   neutral & negative \\
 &  &  & Words & Len. & Len. &  &  &  &  \\
\midrule
\multicolumn{10}{c}{source language} \\
\hline \hline
Slovenian & 10427 & 2017.36 & 349.93 & 4.89 & 23.90 & 152653 & 15.97& 52.03 & 32.00  \\
\midrule
\multicolumn{10}{c}{target languages} \\
\hline \hline
Croatian & 2025 & 1340.29 & 273.18 & 4.14 & 29.11 & 19001 & 14.96  & 63.36 & 21.68 \\ 

Bosnian & 200 & 1326.72 & 222.51 & 4.95 & 25.34 & 1756 & 34.00 & 35.00 & 30.50  \\
Estonian & 100 & 2587.76 & 394.42 & 5.71 & 22.02 & 1791 & 25.00  & 40.00  & 35.00\\
Macedonian & 198 & 2808.70 & 482.43 & 4.88 & 29.47 & 3242 & 20.20 & 60.10 & 19.70  \\
Albanian & 200 & 2988.41 & 543.26 & 4.55 & 29.18 & 3724 & 20.00 & 60.00 & 20.00  \\
Serbian & 200 & 1348.33 & 232.27 & 4.86 & 25.84 & 1798 & 33.00 & 34.50 & 32.50  \\

\bottomrule
\end{tabular}}
    \caption{Dataset statistics.}
    \label{tab:data}
\end{table}
\subsection{Dataset creation}
\label{seubs:data_creation}

The Slovenian dataset was collected by \cite{buvcar2018annotated} and results from an extensive annotation campaign by several annotators. Each document was annotated either as positive, neutral or negative from the readers' perspective. The annotators were asked to answer, ``Did this news evoke positive/neutral/negative feelings?''. Six annotators annotated the dataset; between 2 and 6 annotators annotated each article. The articles were annotated using a five-level Likert scale (1—very negative, 2—negative, 3—neutral, 4—positive, and 5—very positive). The final sentiment of an instance was defined as the average of the sentiment scores given by the different annotators.
An instance was labelled as negative if the average of given scores was less than or equal to 2.4; neutral if the average was between 2.4 and 3.6; or positive if the average of given scores was %more significant %MP: "significant" is definitely not the right word here, beware
greater than or equal to 3.6. The inter-annotator agreement on the document level annotations, as measured by Krippendorf Alpha, is 0.454. In addition, the annotations contain also paragraph- and sentence-level sentiment annotations. While we predict the document-level annotations, the paragraph-level annotations were used for intermediate training in the work of \cite{pelicon2020zero} and also in our introduced POA method.

The Croatian dataset \cite{pelicon2020zero} was collected from 24sata, one of the leading media companies in Croatia with the highest circulation. The annotation followed the same procedure as for the Slovenian dataset. Each article was annotated by 2 to 6 annotators. The agreement, as measured with Krippendorf Alpha, is 0.527.

We have gathered news datasets to test models on news sentiment analysis in a cross-lingual setting for other languages. The novel datasets contain news in five languages: Estonian, Serbian, Bosnian, Macedonian and Albanian. 

We randomly selected 100 articles from the Estonian media house Ekspress Media for the Estonian part of the dataset. Two annotators then annotated the articles. The annotators did not agree in 45 cases. These cases were solved by a third annotator who decided between the two chosen options. The inter-annotator agreement, as computed using Krippendorf Alpha, is 0.335 for the two annotators. 
%\mpu{that's quite low - do we need to say something about that?}
%\hl{andraz, a imamo to kje releasano na clarinu? A je nas EMBEDDIA report public, ce se ti da mi kopiraj link? } \ap{na EMBEDDIA strani imamo objavljen report 4.1, ampak ne omenjamo estonskega dataseta, tako da pomojem še ni public. http://embeddia.eu/outputs/}

For the Serbian, Bosnian, Macedonian and Kosovo news, a Slovenian media monitoring company collected and annotated the news articles. The company manually annotated about 200 articles per language from various topics. A single annotator performed the annotation following the same classes (positive, negative or neutral) as for the datasets above. %\mpu{maybe justify the use of a single annotator? e.g. `as the annotation procedure had been established as reliable, a single annotator [etc]'?}

%We represent the statistical analysis of the datasets in Table \ref{tab:data}. \hl{Boshko, check these parts, I do not see Slovenaian as a total outlier, please rephrase} We notice a minor differences between Slovenian (which in our experiments serves as the source language) and other languages (denoted as target languages) in terms of: average document length, number of words on average and average sentence lengths. Label distribution-wise, we notice a considerable discrepancy between Slovenian and other languages. 

\section{Methodology}

In Section~\ref{sec:related_work},
%MP: it's worth attaching "Section", "Table", "Figure" etc to their reference numbers using latex's non-breaking space ~ as this makes sure you don't get line breaks in the middle
we discussed that one can perform zero-shot cross-lingual sentiment classification in multiple ways. Namely, one can do zero-shot cross-lingual classification by employing a pre-trained multilingual language model; we use this method as a baseline, details in Section~\ref{sec:base_xlmr}. \citet{app10175993} showed that learning paragraph level sentiment-detection before learning document-level classification, as an intermediate task, enhanced the results and offered better classification. Motivated by this, we introduce a novel intermediate-training task (see Section~\ref{inte}). On a parallel note, one of the established ways to solve zero-shot cross-lingual classification consisted of translating the document from the target language (e.g. Slovenian) to the source language (usually English) and performing the classification in the source language. In our experiments, therefore, we use XLMR \cite{conneau-etal-2020-unsupervised} as a baseline model, in combination with different intermediate training strategies (including our new method) and with or without translation, and compare these methods to a strong Large Language Model, based on in-context learning with Mistral7B \cite{jiang2023mistral}.

\subsection{Formulations}
Let $\mathbf{X}$ denote a set of documents %from 
which we learn to classify %them 
into %a 
three classes of sentiment $ y \in \{ \text{positive, neutral, negative} \} $. Throughout our experiments, we will use Slovenian as the source language on which we will learn, so we will use $\mathbf{X}$ and $\mathbf{X^{Slovenian}}$ interchangeably. Formally, we aim to learn a function $ f $ that maps the texts in $\mathbf{X}$ to the correct classes $y$. We characterize the function $ f $ with parameters $\theta$ -- a neural network. Following the deep learning paradigm \cite{lecun2015deep}, we seek to learn the function $ f(\theta, \mathbf{X}) $ such that we minimize the error $\mathcal{L}(\theta, \mathbf{X}, y)$. In our case, we use XLM-Roberta as the base parameterization $\theta_{\text{base}}$ of this function. We define the loss function $\mathcal{L}(\theta, \mathbf{X}, y)$ as categorical cross-entropy, given by:
\[
\mathcal{L}_{sentiment}(\theta, \mathbf{X}, y) = -\sum_{i=0}^{N} \sum_{c=0}^{C} y_{i,c} \log(\hat{y}_{i,c}),
\]
where $N$ is the number of samples, $C$ is the number of classes, $y_{i,c}$ is a binary indicator (0 or 1) if class label $c$ is the correct classification for sample $i$, and $\hat{y}_{i,c}$ is the predicted probability that sample $i$ belongs to class $c$. To learn the parameters $\theta$, we aim to minimize the loss function $\mathcal{L}(\theta, \mathbf{X}, y)$. This optimization problem can be expressed as:
\[
\theta^{*} = \arg \min_{\theta} \mathcal{L}(\theta, \mathbf{X}, y),
\]
where $\theta^{*}$ denotes the parameters after an optimizer of choice minimizes the loss, updating the initial $\theta$. We denote this process as fine-tuning the base XLMR model to the task of document-level sentiment prediction. Since we are interested in the zero-shot cross-lingual setting, our goal is to obtain a representation of $\theta^{*}$, such that we will be able to successfully classify on new $X'$, to get predictions in that target language.

%We consider the translation approach as baseline, described in \ref{seubs:translation} 

%As discussed, cross-lingual sentiment classificaiton can be performed from multiple angles. One can evaluate 
%We evaluate a set of methods in a monolingual setting, as well as in cross-lingual zero-shot  evaluation setting. Additionally, we propose a new paragraph-level intermediate training task, aimed at improving news sentiment classification. In our experiments we use the XLMR model \cite{conneau-etal-2020-unsupervised}, in combination with different intermediate training strategies and with or without translation, and compare these methods also to in-context learning with Mistral7B \cite{jiang2023mistral}.

\subsection{Base model-XLMR}
\label{sec:base_xlmr}

This approach consists of fine-tuning a model on a dataset in a particular language (in our case Slovenian) and then using this model to predict sentiment on a different dataset, either on a test set in the same language (monolingual evaluation) or a or on a dataset in a different language in zero-shot cross-lingual setting. %Because of its proven capability across various language domains, we used the XLMR model. 
For this reason, we select the XLMR \cite{conneau-etal-2020-unsupervised} model as a baseline, as it supports %ed 
all of the languages in our dataset, enabling %us 
efficient transfer in a cross-lingual setting. In this scenario, we utilise the above defined approach of optimising the weights to solve the sentiment classification, where we adapt the XLMR embeddings to the given task. %
XLMR's input is limited to 512 tokens; we follow the original approach on news sentiment classification by \cite{app10175993} for articles exceeding 512 tokens, retaining only the first 256 tokens and last 256 tokens and concatenating them. %ed them for long document presentation. 
In this case we learn in Slovenian: $\theta^* = \arg \min_{\theta} \mathcal{L}(\theta, \mathbf{X, y})$ and use the learned weights to infer predictions in all other languages. This model, without any intermediate training (see methods below), is denoted in our tables \ref{tab:mono_res} and \ref{tab:cross_res} as XLMR (no int. training).

%This approach takes into consideration the fact that for less resourced languages, broader computational resources are needed to achieve a comparable result than the one obtained on well resourced languages such as English and Spanish, since there might be a lack of sufficient enough collection of data for model to be trained on  (Cite AACL article). 

%Thus this approach redirects computational intensity to translating the entire dataset into one of these languages and performing Sentiment Analysis on this augmented dataset, hoping to tackle the problem less resources languages face. 

 %We select this as one additional baseline denoted as \textbf{translated XLMR}.

\subsection{Intermediate training strategies}
\label{inte}

Intermediate training involves adapting the initial XLMR ($\theta_{\text{base}}$) weights on a given intermediate task, which are subsequently used to solve the main task—in our case, learning to classify the sentiment $\mathcal{L}_{\text{sentiment}}$. In this work, we focus on learning to solve paragraph-level sentiment prediction as the intermediate task. Following the methodology introduced in \cite{app10175993}, we denote the paragraph dataset as $\mathbf{P}$ and their corresponding labels $y_p$. We explore two different paragraph-level intermediate training paradigms: PSE proposed by \cite{pelicon2020zero} and a \textit{novel} proposed method POA. 

%(see Section XXX \mpu{which?} for explanation)

\subsubsection{PSE - Paragraph Sentiment Enrichment}
\label{sec:pse}
%We implemented an intermediate training step using paragraph-level sentiment annotations in the Slovenian dataset. We use the approach presented in \cite{app10175993}, where masked language modelling is combined with sentiment prediction for each paragraph as an intermediate training step before the model is fine-tuned again for item-level sentiment prediction. 

%Similar to the base model (see section \ref{sec:base_xlmr}), if a paragraph contained more than 512 tokens, we kept the first 256 tokens and the last 256 tokens of a paragraph. We concatenated them while preserving the sentiment value of the paragraph. 

We denote the method proposed by %Pelicon et al. 
\citet{pelicon2020zero} as Paragraph Sentiment Enrichment (PSE). The idea is based on training the base weights of a model on the task of paragraph-level sentiment prediction and then transferring this knowledge to document-level prediction. Given that for each paragraph \( p \) we have its corresponding label \( y_{p} \), the approach suggests learning to classify the paragraph-level sentiment as the first objective, denoted as \( \mathcal{L}_{\text{paragraph-sentiment}} \). Alongside this, the authors propose adapting to the language via self-supervised masked language modeling of the paragraphs \( \mathcal{L}_{\text{MLM}} \). %The combined objective of this part is \( \mathcal{L}_{\text{paragraph-sentiment}} + \mathcal{L}_{\text{MLM}} \).

The loss for this intermediate sentiment learning is given by:
\[
\mathcal{L}_{\text{PSE}}(\theta, \mathbf{P}, yp) = \underbrace{\mathcal{L}_{\text{paragraph-sentiment}}}_{\text{Paragraph sentiment prediction loss}} + \underbrace{\mathcal{L}_{\text{MLM}}}_{\text{Masked language modeling loss}}
\]
where
\[
\mathcal{L}_{\text{paragraph-sentiment}}(\theta, \mathbf{P}, yp) = -\sum_{p \in \mathbf{P}} \sum_{c=1}^{C} yp_{p,c} \log(\hat{yp}_{p,c}),
\]
is the task of predicting the sentiment at the paragraph level, and
\[
\mathcal{L}_{\text{MLM}}(\theta, \mathbf{P}, \mathbf{P}^*) = -\sum_{p \in \mathbf{P}} \sum_{k=1}^{M} \log P(p_{k} | p^{*}; \theta),
\]
is the self-supervised task of masked language modeling, where we task the model to reconstruct the input (in our case the paragraphs) after self-corruption. Here, \( N \) is the number of paragraphs, \( y_{p_i} \) is the sentiment label for paragraph \( p_i \), \( p_i \) is the input paragraph, \( C \) is the number of sentiment classes, \( M \) is the number of masked tokens in each paragraph, \( p_{j,k} \) is the original token in paragraph \( j \), and \( p_{j}^{*} \) is the masked input tokens in paragraph \( j \).

Once we have optimized this, the initial XLM-R weights (\( \theta_{\text{base}} \)) are transformed to the set of intermediate weights \( \theta_{\text{PSE}} \), which we later use to solve the main objective by fitting the model to the document-level sentiment prediction:

\[
\theta_{\text{PSE}^{*}} = \arg \min_{\theta_{\text{PSE}}} \mathcal{L}_{\text{Sentiment}}(\theta_{\text{PSE}}, \mathbf{X}, \mathbf{y}).
\]

%%\mathcal{L}_{Sentiment}(\theta_{PSE}, \mathbf{X}, y) = -\sum_{i=0}^{N} \sum_{c=0}^{C} y_{i,c} \log(\hat{y}_{i,c}),

\subsubsection{POA - Part Of Article}
\label{poa}
Next, we present our novel approach to the intermediate training -- %the 
Part Of Article (POA). As the name denotes, we aim to learn where a paragraph is positioned within a document as an additional task. The approach modifies the paragraph-level Slovenian dataset by adding an extra label indicating the paragraph's relative position within its article (document). First, we annotate each paragraph of the dataset with its own unique identifier (ID) as a position of the paragraph in the document from which it originated. For instance, the first paragraph contained an ID of 0, the second one received an ID of 1, and so on. We now propose calculating the POA label as:
\begin{equation}
POA(paragraph) = 3 \lfloor \frac{ID(paragraph)}{L(D) - 1} \rfloor
\end{equation}
where $paragraph$ is in the article (document) $D$, $L(D) $ denotes the number of paragraphs in the article(document) $D$, and $\lfloor x \rfloor$ denotes the most extensive whole number, which is %smaller 
less than or equal to $x$. However, we make an exception for the final paragraph in a document, assigning a POA label of 2. 

In summary, each paragraph will be labelled based on its position in the article's beginning, intermediate or last third. The intermediate training step of this approach extends the PSE approach of masked language modelling and paragraph sentiment prediction by introducing the task of predicting the paragraph's relative positional information. The additional information provided, combined with sentiment and the paragraph's position, provides additional information on news sentiment to the model. For example, introductions to news articles typically show more sentiment information than the part describing the backstory of some event. This way, the model could consider these factors and output a more robust sentiment classifier. In addition, POA does not require any additional manual labelling. This enables us to apply it to practically any long-form sentiment analysis approach by splitting our document into smaller fragments, enumerating them, and then computing their POA with the above-mentioned formula. 

Having defined %the 
POA, we next define the POA loss, which consists of the prediction of the correct POA for each paragraph. Formally, for each $paragraph \in \mathbf{P}$, we end up with a class $\text{POA}(\text{paragraph})$. One can quickly realize that this is a classification task that we will introduce to our model, and we end up with a loss:
\[
\mathcal{L}_{\text{POA}}(\theta, \mathbf{P}, y^{\text{POA}}) = -\sum_{i=1}^{N} \sum_{c=1}^{C} y^{\text{POA}}_{i,c} \log(\hat{y}^{\text{POA}}_{i,c}),
\]
where $y^{\text{POA}}$ is the true label for the POA classification and $\hat{y}^{\text{POA}}$ is the predicted probability for the POA classification.

Finally, the final loss of this \textit{novel} intermediate training task is denoted as:
\[
\underbrace{\mathcal{L}_{\text{paragraph-sentiment}} + \mathcal{L}_{\text{MLM}}}_{\text{PSE loss (Section 3.2)}} + \underbrace{\mathcal{L}_{\text{POA}}}_{\textit{novel} \text{ Part Of Article loss}}.
\]

\begin{figure}[H]
    \centering
    \includegraphics[width=\textwidth]{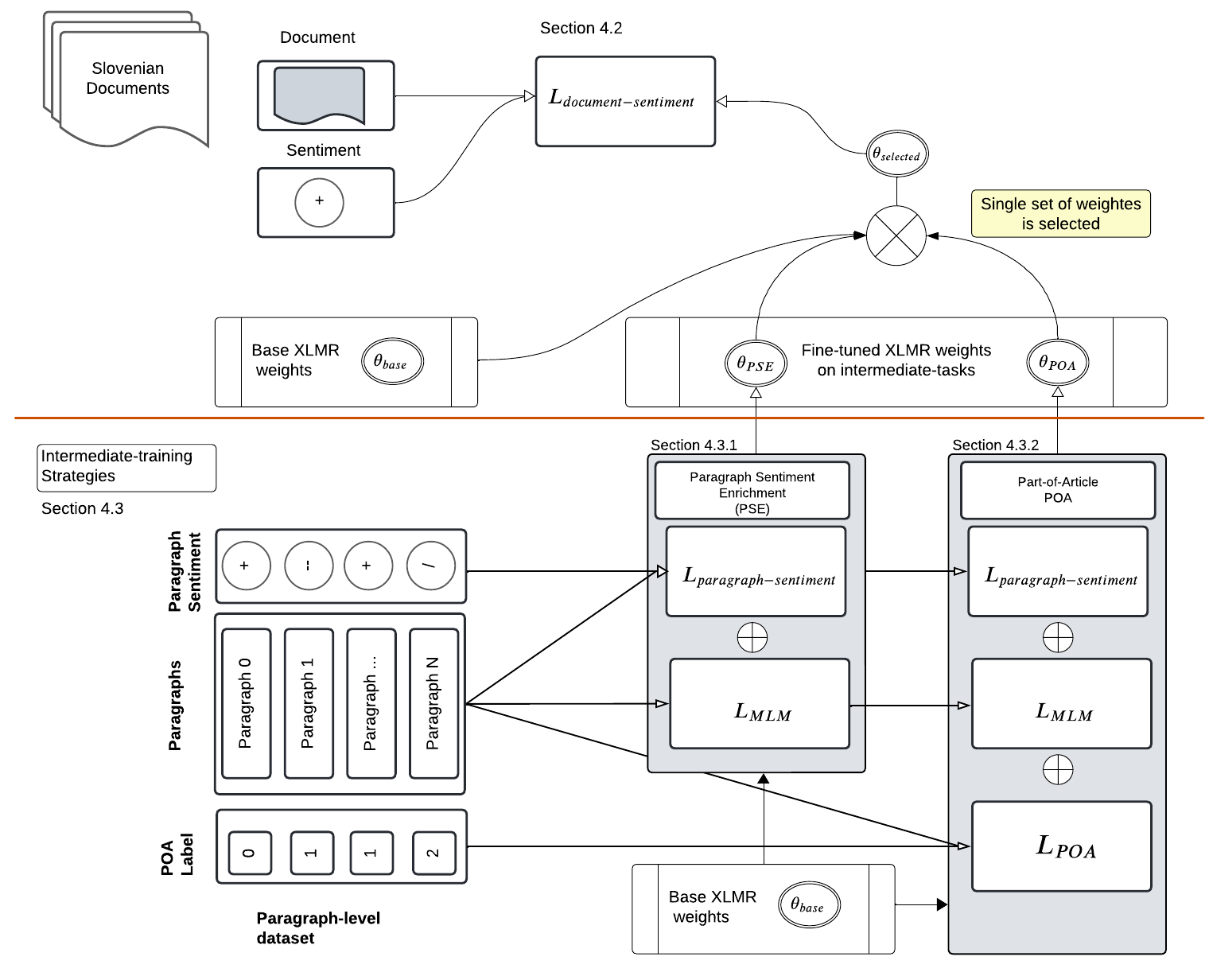}
    \caption{Diagram illustrating the selected and proposed training strategies. The first approach, PSE (Section 4.3.1), is based on paragraph-level sentiment analysis. The second approach, POA (Section 4.3.2), introduces the newly proposed method that includes additional information about sentiment positions. Alternatively, the diagram shows that one can entirely skip the intermediate training phase and focus solely on fine-tuning the base XLMR weights for the main task, which in this case is document-level sentiment training in Slovenian.}
    \label{fig:diagram-label}
\end{figure}

After optimizing this loss, the base $\theta_{\text{base}}$ weights converge to $\theta_{\text{POA}^{*}}$, which we utilize to initialize the the main task of document-level sentiment classification. We show the diagram of fine-tuning based approaches in Figure~\ref{fig:diagram-label}.

%:

%\[
%\mathcal{L}_{\text{Sentiment}}(\theta_{\text{POA}^{*}}, \mathbf{X}, y) = -\sum_{i=1}^{N} \sum_{c=1}^{C} y_{i,c} \log(\hat{y}_{i,c}),
%\].

%\subsubsection{Translation into English}
\subsection{Translation strategies -- from source to English}
\label{seubs:translation}

While multilingual models, such as XLMR, support a large number of languages, their performance might be lower for less-resourced languages compared to the dominant languages \cite{bapna-firat-2019-simple,pfeiffer-etal-2020-mad,winata2022cross}. The main reason for poor performance is the model's inability to equally represent all languages in its vocabulary and representation space. Translating the data from less-resourced language (like Macedonian or Albanian for example) to well-resourced one like English, can benefit classification tasks \cite{unanue2023t3l}. To assess the benefits from translation, we leverage the recently proposed No Language Left Behind model (NLLB) \cite{nllbteam2022language} and compare the performance of the cross-lingual transfer when using the original languages, compared to using all the data in English. We consider evaluating the base and the two intermediate training strategies, both with and without translation. The setting where we train the XLMR model with translation is denoted as \textit{Translation Yes} or \textit{No}.

\subsection{In-context learning}
The in-context learning (ICL)  approach \cite{dong2023survey} capitalizes on a large language model (LLM)'s capacity for analogical reasoning. Distinct from many machine learning paradigms, ICL does not involve a training phase where the model's parameters are updated through gradient back-propagation. Instead, the pretrained model's weights are frozen, and a straightforward protocol is employed to elicit the model's output. This process involves gathering data in its natural language form and restructuring it into a standardized template. For example, the template for a single article example might be structured as follows: ``Article >> (Article's contents) <<, sentiment: (Article's sentiment)''. 
The input to the model is then a single example or a concatenation of a few examples, followed by an incomplete example where the article's sentiment is missing and must be predicted by the LLM. 

The query is introduced with the system prompt, to provide the model with a general context of the problem and to specify the desired format of the model's output. In our experiments, we provided the following instructions:

\textit{Welcome to your role as a Sentiment Analyst. Your job is crucial and involves meticulously reading and understanding the text provided to determine its overall sentiment. It's important to approach this task with attention to detail, considering the subtleties of language, tone, context, and the intent behind the words.}

\textit{As a Sentiment Analyst, your primary task today is to analyze the following news article.
After a thorough review, classify the sentiment of the article as "positive," "negative," or "neutral."
Remember, your analysis must be based on concrete evidence from the text.
As an output you need to only provide the sentiment, so output "positive" if the sentiment is positive, "neutral" if the sentiment is neutral, and "negative" if the sentiment is negative.}

The in-context learning approach was conducted with the Mistral7B model \cite{jiang2023mistral}. In all of the ICL experiments, the system prompt given above was used as the beginning of the query. We investigated various \textbf{in-context learning} (ICL) configurations, specifically zero-shot, one-shot, and three-shot settings. 
The zero-shot approach consists solely of a query that provides the model with context and instructions, without including any examples. The one-shot setting incorporates one example for each sentiment label (i.e.\ three in total); %Conversely, 
the three-shot setting includes three examples for each sentiment label.% In both the one-shot and three-shot configurations, the previously mentioned system prompt is added to the input query to guide the model's response. \mpu{I would remove the last sentence, it's basically the third time we've said it in the last couple of paragraphs.}

The three shot setting consistently achieved the highest performance score across all languages; in the results of Section~\ref{sec:results}, we report three-shot results.  

The system prompt and the input examples were always translated to English, as the model has seen far greater data in English.
For the evaluation examples, on the other hand, we compared the setting with and without translation (keeping the articles in original text and comparing it with their translations in English).

In the non-translation approach, no text translation was performed, requiring the LLM to process multiple languages, including some written in non-Latin scripts such as Macedonian, which uses the Cyrillic alphabet. Conversely, in the translation approach, we translated the entire corpus into English and supplied the translated data to the model, restoring a monolingual setting. 

We did however provide the query to the model in English in both cases -- motivated by the results of Qin et al. \cite{qin-etal-2023-cross}. In the one-shot and three-shot setting we also provided the model the example articles in English. The translation and non-translation approach thus only differs in input to the model and nothing else. %\mpu{this seems to partially duplicate what we've already said two paras above}

%In the case of translation we denote the learning of base models as  $\theta^* = \arg \min_{\theta} \mathcal{L}(\theta, \mathbf{X^{Slovenian -> English}}, y^{Slovenian})$ 

\section{Experimental setting}
We are interested in evaluating the zero-shot transfer capabilities of the base approach (Section \ref{sec:base_xlmr}), as well as the intermediate-training PSE and novel POA approaches (Section \ref{sec:pse} and Section \ref{poa}) alongside the in-context learning on translated and non-translated data. The main research questions we address are:
\begin{itemize}
    \item \textbf{RQ1:} How does the novel intermediate training POA approach perform in monolingual and zero-shot cross-lingual settings for document-level sentiment classification?
    \item \textbf{RQ2:} What is the best method for cross-lingual sentiment classification in the era of in-context learning?
    \item \textbf{RQ3:} How does translating from a less-resourced (e.g. Slovenian) to high resourced language    (English) affect the cross-lingual transfer performance across different methods? %\mpu{maybe better to use an open wh-question e.g. "how does translating [...] affect performance"}
    %\item Is language similarity preserved under various training paradigms?
    \item \textbf{RQ4:} How can we explain potential discrepancies between source and target languages, with optimal-transport dataset distance and topic modeling?  
\end{itemize}

In all settings we start from Slovenian as source language, and we transfer to Croatian, Bosnian, Estonian, Macedonian, Albanian and Serbian. %In the case of 
We also experiment with monolingual (i.e.\ Slovenian) evaluation, to help investigate which strategy is optimal for training when auxiliary paragraph-level data is available. Next, we explain how we prepare the data for our training and testing languages, as well as how we evaluate and specify our experiments.

\subsection{Data splits}

The intermediate training used the same training, validation and test splits as the base model approach. 

\subsubsection{Splitting the Slovenian dataset}
%\mpu{`Slovenian' here vs. `Slovenian' in most other places?}}

The procedure for splitting the data in the Basic approach and Basic + Translation approach is the standard train, validation and test split. We employ 10 fold validation, so we hold out 10\% of the data for testing purposes, while splitting the remaining 90\% of the data into a train and validation split (80\% and 20\%).

In the approaches with intermediate training, namely PSE and POA, we are considering paragraph-level information. In these settings, we keep the test set (used for document level evaluation in the Basic approach) untouched, while paragraph-level pretraining uses 10-fold cross-validation on the training corpus. This ensures that paragraphs used in the final evaluation step are excluded from pre-training, to prevent data leakage.

In the zero-shot in-context learning (ICL) approach, we infer the results for the entire dataset, but report the F1-score results across 10 folds (%sames as for 
as with the trained models) allowing to compute also standard deviation.

We %omitted the examples 
exclude the examples that appear in the %example 
prompt from the evaluation set. Because the number of these examples (9 in total in the three-shot setting on a large Slovenian dataset of 10 427 articles) is %marginal 
so small in comparison to the dataset size, results are still comparable across methods.

\subsubsection{Splitting other datasets}
For other languages than Slovenian, no splitting was needed as we evaluated the models in a zero-shot setting. We did however run 10 iterations in the basic, PSA and POA approaches. This is because in all of these methods, the models were trained on different data splits because of the 10-fold cross validation approach. We report the average of these results for each approach respectively. In the ICL approach no splitting of data was needed. %\mpu{this is a bit confusing - we're saying here that no splitting is needed for ICL, but in the previous paras we said we run 10 folds with ICL, and in the results section we report CIs - we should explain more clearly here} 
In the one-shot and three-shot ICL setting, no risk of data leakage was present as we used examples from the Slovenian dataset.

\subsection{Evaluation metrics}
In all cases, we report the F1 score  and the standard deviation respective of each setting.

\subsection{Experiment specifications}
Finding the optimal values for the $\theta$ parameters is intractable, so we use the gradient-based AdamW optimizer \cite{loshchilov2018fixing} to adapt the $\theta$ weights to the data. The model is trained for 3 epochs with a batch size of 8. We set the learning rate to $2 \times 10^{-5}$, apply a weight decay of 0.01, and use an Adam epsilon of $10^{-8}$. We implement the experiments in PyTorch \cite{paszke2017automatic} and use HuggingFace Transformers \cite{wolf-etal-2020-transformers}. All experiments are executed on a single 32GB V100 GPU.

%In all steps of our experiments involving training a model (with exception of the ICL), we used the following parameters :
%\begin{itemize}
%    \item Epochs : 3
%    \item Batch size : 8
%    \item Learning rate : $2 \cdot 10^{-5}$
%    \item Weight decay : 0.01
%    \item Warm-up proportion : 0.1
%    \item Adam epsilon : $1 \cdot 10^{-8}$
%\end{itemize}
%Notably, we used the AdamW optimizer \textbf{REF} with a linear scheduler with warmup. 

%Monolingual section, where we present results obtained on the Slovenian dataset, and Cross-lingual section, where we present results of the same methods on all other datasets. This is done for multiple reasons ; the Slovenian dataset is much larger and more difficult than the other one, the Monolingual evaluation is not subject to zero-shot evaluation while the same cannot be said for the other case. 

\section{Results}
\label{sec:results}
Here, we present the performance of the proposed intermediate learning method POA in monolingual setting in Subsection \ref{subsec:mono}, followed by the comparison of the selected zero-shot cross-lingual classification methods in Subsection \ref{subsec:zeroshot}. Following that we %do 
provide a
qualitative analysis on the predictability of the performance of methods and qualitative analysis of the datasets.

\subsection{Monolingual evaluation}
\label{subsec:mono}
Table \ref{tab:mono_res} shows the results for the Slovenian dataset. We can see that all but the basic non-translation approach outperform ICL. Furthermore, we find that translation into English improved the baseline approach, while translation did not significantly affect the PSE and POA results. We note that the standard deviation of the baseline approach without translation is very high compared to the other settings, which leads us to conclude that this may not be a reliable result.
We perform a paired T-test across all strategies to assess the statistical significance of the effects of translation in the general setting. We find that there is no statistical significance in influencing the results (t-statistic=-1.15, p-value=0.33). We also find that intermediate training strategies improve performance and robustness when used to initialise the document-level classification weights. We find an increase of 5.54\% with the PSE strategy and 5.62\% with the novel POA strategy when no translation is included. For translation, we found an increase of 0.08\% for PSE and 0.1\% for POA. We attribute this to the fact that when translating the data into English, the models try to explore the same local minima, while the low coverage of Slovenian in the non-translated case gives the model better opportunities to explore the landscape when solving the additional tasks, especially during the MLM phase. One must also consider that in some cases the translation quality might translate the items without conveying the original meaning, which is due to the less-resources of the Slovenian in the original NLLB corpora. Also note that the results are higher compared to previous results of Pelicon \cite{app10175993} on the same test split. They used the mBERT model and achieved 66.33\% F1 score when they used intermediate training, while our XLM-R approach with intermediate training (their PSE setting or our novel POA approach) both achieve results surpass this score, with the POA approach in particular achieving the state-of- the-art result of 71.02\% F1-score. We credit this boost both to the more advanced method to represent the documnets - XLMR and the fact that we perform more advanced intermediate training.

\begin{table}[ht]
    \centering
    \caption{Monolingual evaluation on the Slovenian dataset. Intermediate-training denotes which strategy we use: PSA, POA or none. Translation denotes whether we translate the data or not. For in-context learning, the 3-shot setting is used.}

    \begin{tabular}{|c|c|c|}
     \hline
     \textbf{In-context learning} & \textbf{Translation} & \textbf{Slovenian} \\ \hline
          & No & 65.52 ± 2.69 \\
          & Yes & 66.76 ± 3.13 \\ \hline
       
       \textbf{Intermediate training} & & \\ \hline
          XLMR (no int. training) & No & 65.30 ± 14.07 \\     
          XLMR (no int. training) & Yes & 70.60 ± 3.46 \\
         PSE & No & 70.84 ± 3.09 \\
         PSE & Yes & 70.68 ± 3.04 \\
         POA & No & \textbf{71.02} ± 3.71 \\
         POA & Yes & 70.70 ± 3.36 \\ \hline 
    \end{tabular}
    \label{tab:mono_res}
\end{table}

%hi boshko

\subsection{Zero-shot evaluation}
\label{subsec:zeroshot}

\begin{figure}[h!]
    \centering
    \resizebox{0.9\textwidth
    }{!}{\includegraphics{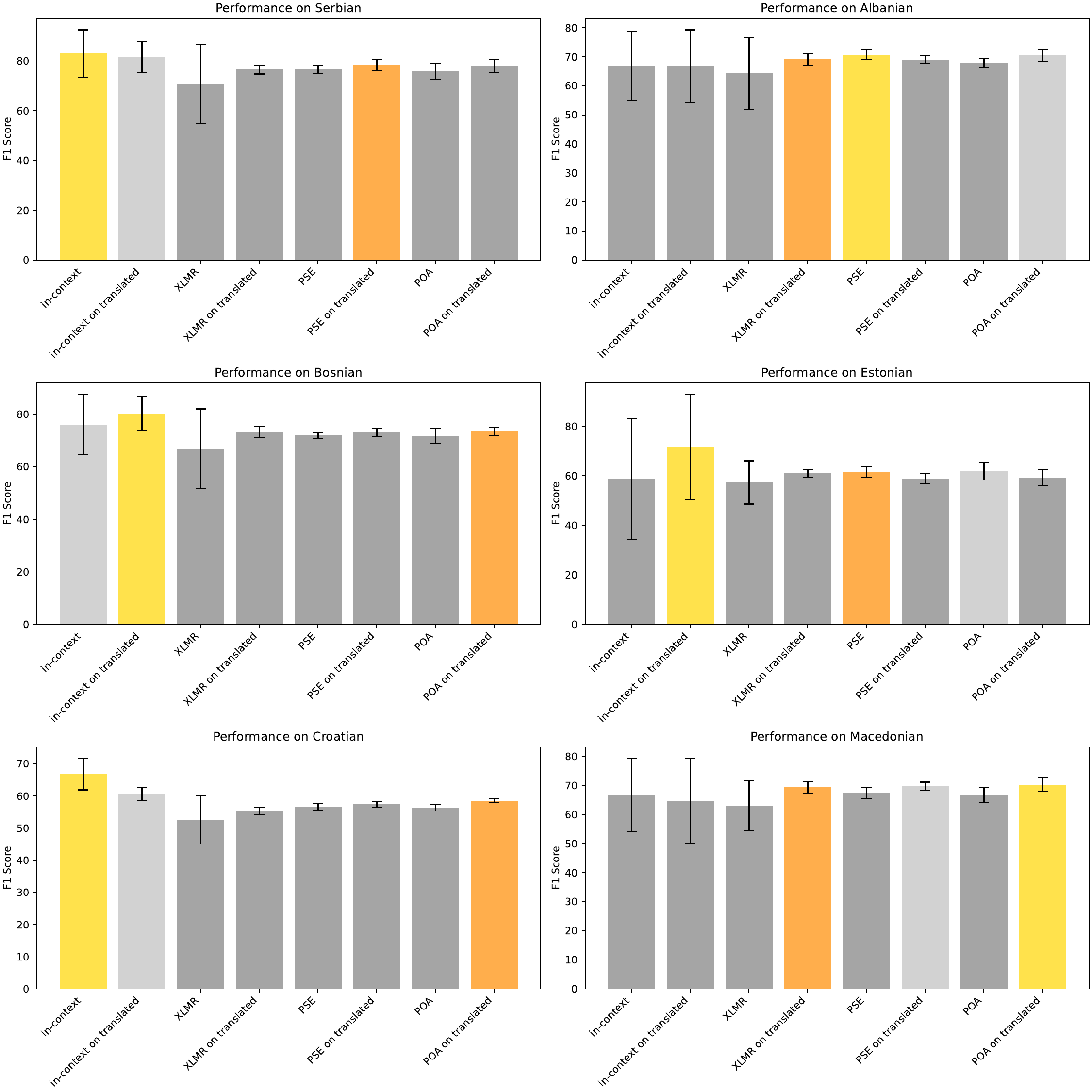}}
    \caption{Representation of language performance of each method. For each language, the gold-coloured bars represent the best method, followed by the second-best in silver and the third-best in bronze.}
    \label{fig:perf_across_langs}
\end{figure}

The results of the zero-shot evaluation on other datasets are shown in Table \ref{tab:cross_res} and visualized in Figure \ref{fig:perf_across_langs}. The results show that in cross-lingual context the in-context learning outperforms most other models. This coupled with the fact that the in-context approach does not require any pretraining or fine-tuning makes it an efficient method for cross-lingual sentiment analysis. On the other hand we do note that for Albanian and Macedonian better results are achieved with intermediate training.

%%%%%%%%%%%%%%%

\begin{table}[H]
    \centering
    \resizebox{\textwidth}{!}{\begin{tabular}{|c|c|c|c|c|c|c|c|}
     \hline
     \textbf{In-context learning} & \textbf{Transl.} & \textbf{Serbian} & \textbf{Albanian} & \textbf{Bosnian} & \textbf{Estonia} & \textbf{Croatian} & \textbf{Macedonian} \\ \hline
          & No & \textbf{83.03} ± 9.48 & 66.83 ± 12.06 & 76.15 ± 11.51 & 58.75 ± 24.44 & \textbf{66.79} ± 4.86 & 66.63 ± 12.57 \\
          & Yes & 81.74 ± 6.25 & 66.78 ± 12.52 & \textbf{80.25} ± 6.54 & \textbf{71.71} ± 21.18 & 60.56 ± 2.04 & 64.64 ± 14.61 \\ \hline
       
       \textbf{Intermediate training} & & & & & & & \\ \hline
          XLMR (no int. training) & No & 70.75 ± 16.00 & 64.39 ± 12.37 & 66.92 ± 15.17 & 57.32 ± 8.71 & 52.69 ± 7.53 & 63.12 ± 8.50 \\     
          XLMR (no int. training) & Yes & 76.57 ± 1.81 & 69.12 ± 2.04 & 73.26 ± 2.13 & 61.00 ± 1.60 & 55.39 ± 1.06 & 69.36 ± 1.92 \\
         PSA & No & 76.70 ± 1.73 & \textbf{70.77} ± 1.70 & 71.96 ± 1.23 & 61.63 ± 2.15 & 56.57 ± 1.08 & 67.47 ± 1.92 \\
         PSA & Yes & 78.41 ± 2.19 & 69.09 ± 1.46 & 73.06 ± 1.67 & 58.96 ± 2.10 & 57.50 ± 0.90 & 69.79 ± 1.38 \\
         POA & No & 75.89 ± 3.09 & 67.81 ± 1.66 & 71.72 ± 2.91 & 61.86 ± 3.51 & 56.34 ± 0.95 & 66.80 ± 2.60 \\
         POA & Yes & 78.03 ± 2.61 & 70.45 ± 2.15 & 73.59 ± 1.62 & 59.30 ± 3.31 & 58.59 ± 0.49 & \textbf{70.33} ± 2.42 \\ \hline
    \end{tabular}}
    \caption{Zero shot evaluation on all datasets . Intermediate-training denotes which strategy we use: PSA, POA or none. Translation denotes whether we translate the data or not.}
    \label{tab:cross_res}
\end{table}

%%%%%%%%%%%%%%%%%%

%\hl{this section should be CLEANED UP - would it be possible to keep only Fig 2., but keep the significan results in bold. Also the names of the models have to be done the same as in the other table (capitalised, for translation you can keep at the end -Trans.}

We also note that the in-context learning results are not stable (high standard deviation), which is also characteristics for trained XLMR model without translation. Interestingly, translation as well as intermediate learning strategies improve the robustness of the model (lower standard deviation).

\subsection{On the predictability of performance of methods}
\label{subs:otdd}
\begin{figure}[H]
 \centering
 \includegraphics[width=0.9\textwidth]{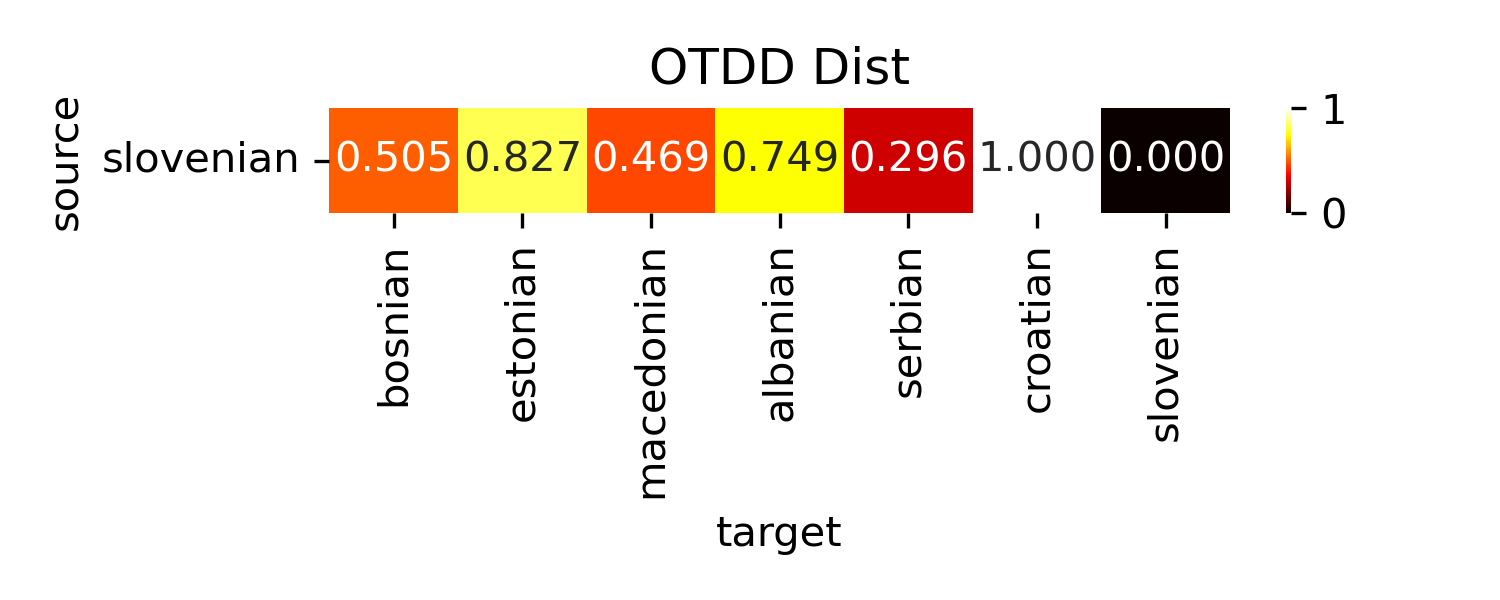}
 \caption{Optimal transport dataset distance between the Slovenian dataset (source) and the target datasets (destination)}
 \label{fig:otdd}
\end{figure}

In Figure \ref{fig:otdd}, we analysed the distance between datasets based on the Optimal Transport Dataset Distance (OTDD) metric \cite{alvarez2020geometric}. We compared the combined Slovene (train and test) with the test data of other languages. 
\vspace{-1em}
The OTDD metric is based on the principles of optimal transport and calculates the coupling between probability masses of labels distributed by examples represented by an encoder in a vector representation. We use the XLMR-based sentence transformer representation\footnote{sentence-transformers/XLMR-distilroberta-base-paraphrase-v1} to encode the datasets presented in \cite{reimers2019sentence}.

\begin{figure}[H]
    \centering
    \resizebox{\textwidth}{!}{0.8\includegraphics{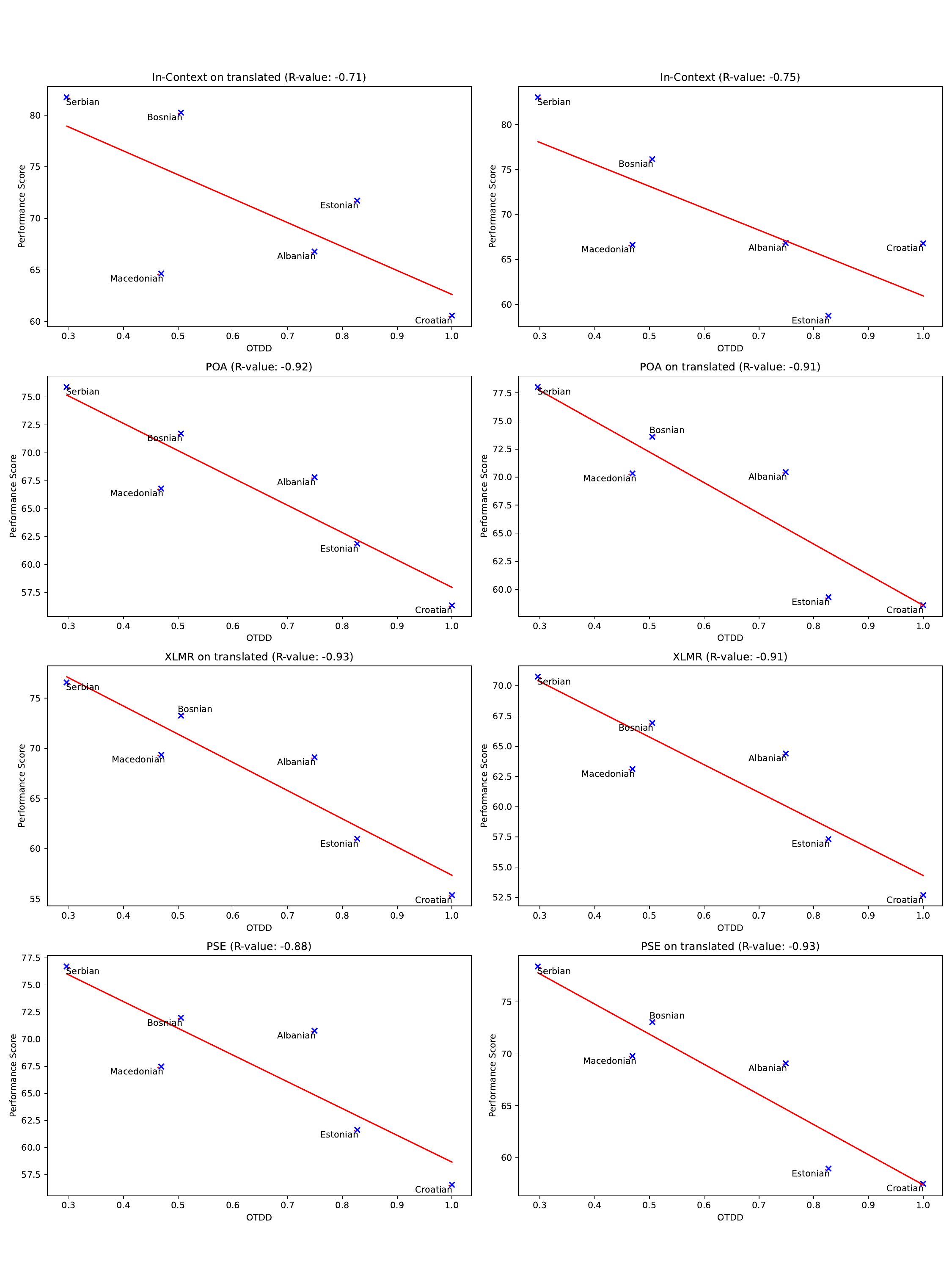}}
    \caption{Correlation of the OTDD metric and the performance of each metric across langauges.}
    \label{fig:corr}
\end{figure}

The results show that the labelled samples from Serbian (normalised distance of 0.30) are closest to Slovenian, followed by Macedonian (normalised distance of 0.47) and Bosnian (normalised distance of 0.51). This is followed by Albanian with a normalised distance of 0.75 and Estonian with 0.83. Unexpectedly, Croatian was the language furthest away from Slovenian in terms of OTDD distance, primarily due to the shift in distributions (recall Table \ref{tab:data}). Figure~\ref{fig:corr} shows the correlations between OTDD distance and performance for the different methods. We observe a consistent negative correlation between OTDD and performance. This is expected, as an increase in OTDD distance corresponds to a decrease in performance. For zero-shot transfer tasks, methods like POA and XLMR-base, while highly performant, are more sensitive to OTDD.  This suggests a need for closer dataset alignment or additional adaptation techniques to mitigate performance drops. Using translated datasets can improve performance for most methods, particularly for POA, XLMR, and PSE. When transferring from Slovenian, languages with lower OTDD values (e.g., Serbian, Macedonian) should be prioritized to maximize model performance.

\subsection{Semi-automatic qualitative analysis}
Next, we conduct a semi-automatic qualitative analysis to examine whether the high OTDD distances can be attributed to topical discrapancies between datasets. We generate topics on the Slovenian dataset using BERTtopic \cite{grootendorst2022bertopic} for generating topics on the Slovenian dataset. Next, we use this model to annotate the articles in the remaining documents. We use the translated documents from Section \ref{seubs:translation} for consistency: related work shows that encoder-based LLMs place similar languages in similar clusters, which might disrupt
our analysis. Following Koloski et al.~\cite{koloski2024aham}, we consider using LLMs to devise more understandable human topics. We use GPT-4 \cite{openai2024gpt4} for generating topics and the XLMR above sentence transformers variant for encoding the articles. We set the maximum number of topics to 30. Since we have discrete topics per language and language label distributions, we consider the $\chi$-square test to assess any statistical differences. As the datasets are unbalanced, we consider sampling within a similar scale (always down-sampling Slovenian) for fair comparison. The resulting 29 specific and one general (outlier) topic are shown in Figure \ref{fig:enter-label}. The most-prevalent (outlier topic) was `finance and governance' with around 30\% (3192) articles. The specific topics were the topics of `Slovenian Banking and Economy', `Stock market Trends', and `Finance and Banking' between 10\% and 15\%. These results do not surprise us, since Economy and Finance are amongs the most used topics according to ~\citeauthor{lah2020journalism}. Following were the topics on `aviation`, `Slovenian Politics' and `unemployment'. The least represented topics were `gender equality in business and technology' and the `corporate awards and mergers', which we entitled to their specificity and regularization of media \cite{drvzanivc2022gap}. The results show that the dataset originates from general news, with few segments of it being fairly specialised. We hypothesize that due to the generality of the dataset, the learned knowledge should transfer cross-lingually.

\begin{figure}[H]
    \centering
    \includegraphics[width=\textwidth]{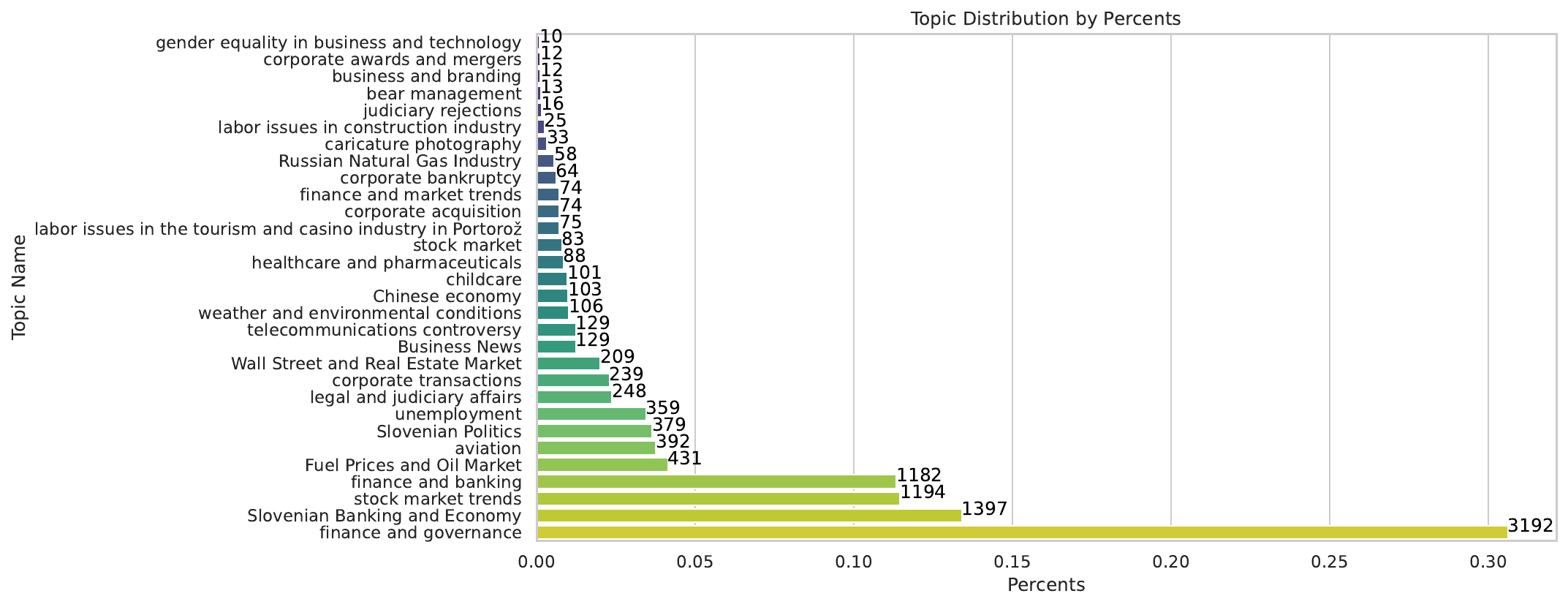}
    \caption{Topic distribution in the original Slovenian datasets.}
    \label{fig:topic-Slovenian}
\end{figure}

%\begin{figure}[H]
%    \centering
 %   \includegraphics[width=\textwidth]{newplot (75).png}
%    \caption{Hierarchical layout of the detected topics.}
 %   \label{fig:enter-label}
%\end{figure}

\begin{figure}[H]
    \centering
    \includegraphics[width=0.80\textwidth]{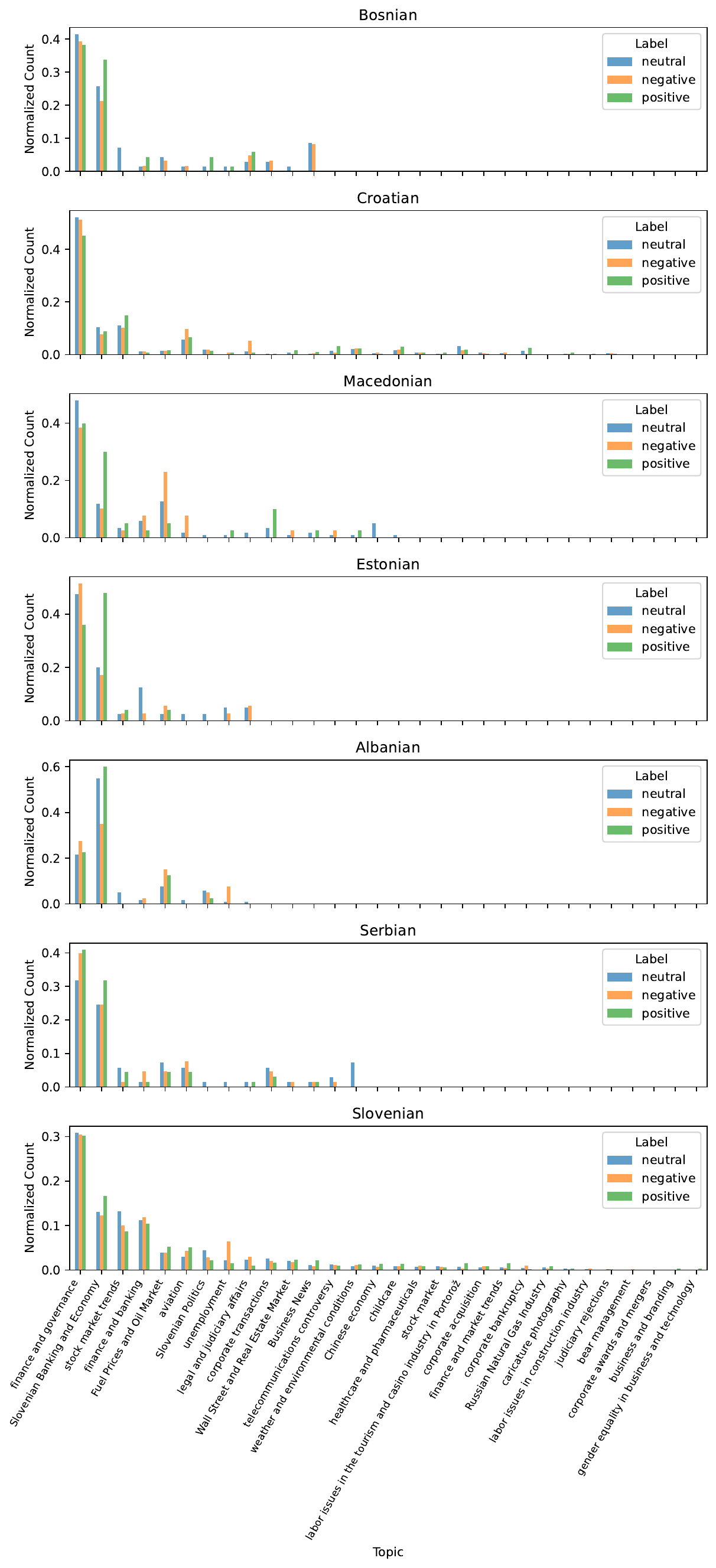}
    \caption{Comparison of topic distribution.}
    \label{fig:Topic-Language}
\end{figure}

We present the distribution of topics per language in Figure \ref{fig:Topic-Language}. Minor shifts are observed for Serbian and Estonian, while more significant discrepancies are noted for Croatian, followed by Macedonian. These observations align with the OTDD results from Section \ref{subs:otdd}. We assess these distribution shifts using the Chi-squared test, as detailed in Table \ref{tab:chi2_top_mod}. The results show that there is a significant difference between Slovenian and Croatian ( $\chi^2$=173.82, p-value<0.01), Slovenian and Macedonian ( $\chi^2$=64.46, p-value=0.02), Slovenian and Albanian ( $\chi^2$=59.85, p-value=0.02) and Slovenian and Serbian ( $\chi^2$=66.193, p-value=0.04).
The results in Figure \ref{fig:Topic-Language} show clear differences in the topic distributions between the Slovenian and Croatian datasets, as evidenced by our detailed topic analysis, which aligns with the OTDD discrepancy. We hypothesize that topic discrepancies between datasets can be used for assessing disstribution shifts. 
Note that these results are based on approximate translations on the NLLB model and should be considered with some variation. However, the indication that Slovenian and Croatian differences can be analysed through this lens, enabling us to better understand the difference in topic distributions and suggest an explanation for the performance discrepancies. We propose that, in future work, topic modeling of datasets by first translating them into English and then analyzing the distribution mismatch could provide an approximation of the transferability of performance across datasets.

\begin{table}[H]
    \centering
    \begin{tabular}{llrr}
    \toprule
    Source & Target Language & Chi2 Statistic & p-value \\
    \midrule
    slovenian & bosnian & 65.484 & 0.070 \\
    slovenian & croatian & \textbf{173.819} & \textbf{0.000} \\
    slovenian & macedonian & \textbf{64.464} &\textbf{ 0.024} \\
    slovenian & estonian & 39.134 & 0.419 \\
    slovenian & albanian & 59.847 &\textbf{ 0.023} \\
    slovenian & serbian & 66.193 & 0.042 \\
    \bottomrule
    \end{tabular}

    \caption{Chi2 topic modeling.}
    \label{tab:chi2_top_mod}
\end{table}

\section{Discussion and Limitation}
The two in-context learning variants (with and without translations) achieved comparable and in most cases the best scores. Given the significant differences in computational requirements between various approaches, in some cases, in-context learning provides a more flexible and high-performing alternative to traditional machine learning methods. It can also address common challenges such as the limitation of fixed input sizes. %The results show our approaches are statistically significant \mpu{an approach can't be "statistically significant". Do you mean "show our approaches give statistically significant performance improvements over the XLMR movel" or similar?} and do indeed improve the performance of the XLMR model in this diverse linguistic landscape. 
\par A notable limitation of our study is the uneven amount of data across languages. The Slovenian language is by far the most widely represented in the corpus, followed by Croatian. Such limitations obscure the direct comparison of performances on these subdatasets, while still allowing us to explore the effectiveness of various approaches. A surprising result is the fact that the performance on the Estonian dataset is not obviously diminished as we had expected since it is the only non-Slavic language in our study. As expected, the results on the Macedonian dataset are lower, owing to the fact it is written in the Cyrillic alphabet. An improvement on these limitations would be acquiring a dataset with a comparable amount of labeled data across languages and expanding the evaluation phase to other European and even non-European languages. 
The POA approach achieved the highest performance on the Slovenian dataset, while not showing direct improvements across all other languages. The explanation behind this may be the fact that news articles from other datasets had an inherently different document structure than the articles in the Slovenian dataset, the relative position information conveyed in particular document sections may differ across different media houses. This difference may be even more amplified in our case, as our data involves changes not only in media house, but across country and culture. Another shortcoming of the POA approach may stem from an assumption that an article's structure may be described linearly with only 1 parameter. This leads us to further research into the correlation between the surface structure of a news article to its semantic information. 

We conclude by returning to our research questions:
\begin{itemize}
    \item \textbf{RQ1:} How does the novel intermediate training POA approach perform in monolingual and zero-shot cross-lingual settings for document-level sentiment classification? \\
    \textbf{ANSWER:} The novel POA approach outperforms all methods in the monolingual setting and performs competitively 
    against other approaches, including in-context learning.
    \item \textbf{RQ2:} What is the best method for cross-lingual sentiment classification in the era of in-context learning? \\
    \textbf{ANSWER:} The in-context learning method performs remarkably well. However, due to its power-hungry requirements, the intermediate training and translation-boosted methods are viable approaches, especially when on-demand labeling of documents is needed, such as in the news industry.
    \item \textbf{RQ3:} Does translating from less-resourced (e.g., Slovenian) to high-resourced language (English) always positively affect the cross-lingual transfer performance across different methods? \\
    \textbf{ANSWER:} We find that there is no golden rule, and this varies across methods and languages. This is likely due to the translation model capacity and language coverage. We propose investigating in the future how different translation models affect this perspective.
    \item \textbf{RQ4.} How can we explain potential discrepancies between source and target languages, with optimal-transport dataset distance and topic modeling?   \\
    \textbf{ANSWER:} We find that datasets that are semantically distant in terms of labels, context, and topic distributions tend to transfer knowledge less effectively, resulting in poorer performance. %\mpu{you could add RQ4 about distance between languages (``does cross-lingual transfer perform better between more similar languages?''), concluding that it's more about the distance (OTDD) between datasets ... ?}
\end{itemize}

\section{Conclusion}
In this paper, we explore various techniques for cross-lingual sentiment analysis. We introduce a novel intermediate step, POA, designed specifically to improve sentiment analysis on long documents. We compare this approach with other standard techniques, such as zero-shot and translation. We demonstrate the benefits and limitations of POA and its impact on the stability and robustness of the model across this diverse linguistic environments using the XLMR model. Additionally we tested the in-context learning approach with LLMs and achieved pleasing results. We may benefit from gathering more data in the languages which are in our case less abundant. Additionally this leads to delve further in the core structure of news articles and their content across different languages and cultures, looking for invariant patterns, features and consistencies.

\section*{Acknowledgments}

The authors acknowledge financial support from the Slovenian Research and Innovation Agency through research core funding (No. P2-0103) and projects No. J4-4555, J5-3102, L2-50070, and PR-12394.
\bibliography{sample}

\end{document}